\documentclass[journal,twoside,web]{ieeecolor}
\usepackage{generic}
\usepackage{cite}
\usepackage{amsmath,amssymb,amsfonts}
\usepackage{algorithmic}
\usepackage{graphicx}
\usepackage{textcomp}
\usepackage{xcolor}
\usepackage{url}
\usepackage{hyperref}
\graphicspath{{figures/}}
\def\BibTeX{{\rm B\kern-.05em{\sc i\kern-.025em b}\kern-.08em
    T\kern-.1667em\lower.7ex\hbox{E}\kern-.125emX}}
\begin{document}
\title{EveLoad: Cognitive Workload Recognition from Event-Based Eye Movements}

\author{Guorui Lu*, Shaohua Guan*, Zhen Xu, Qinyu Chen
\thanks{* Guorui Lu and Shaohua Guan contribute equally.}
\thanks{Guorui Lu, Shaohua Guan, Zhen Xu, Qinyu Chen are with Leiden Institute of Advanced Computer Science (LIACS), Leiden University, The Netherlands. (g.lu@liacs.leidenuniv.nl, q.chen@liacs.leidenuniv.nl)}
}

\maketitle

\begin{abstract}
Cognitive workload monitoring is important for adaptive rehabilitation and assistive interfaces, where task difficulty, pacing, and feedback should be adjusted according to the user's cognitive state to avoid overload and under-challenge.
Emerging extended reality (XR) and robot-assisted rehabilitation environments provide controllable training tasks, but they require unobtrusive sensing methods that can capture rapid ocular dynamics during interaction.
Existing eye-movement-based cognitive workload recognition methods mainly rely on frame-based eye trackers, which often suffer from limited temporal resolution and degraded robustness under rapid eye movements.
In contrast, event cameras provide microsecond-level temporal resolution, high dynamic range and low latency, making them suitable for capturing fine-grained ocular dynamics.
Many previous studies rely on free-viewing or similar paradigms, where gaze locations can vary across tasks. As a result, models may learn associations between gaze-location distributions and cognitive workload, rather than workload-related eye movement characteristics themselves.
In this work, we introduce EveLoad, which, to the best of our knowledge, is the first event-based eye-movement dataset~\footnote{The dataset has been publicly released and is available on \href{https://ieee-dataport.org/documents/eveload-cognitive-workload-recognition-eye-movements-using-event-cameras}{IEEE DataPort}.} with graded cognitive workload annotations, collected from 20 healthy participants under spatially constrained and task-driven conditions using a controlled N-back-guided fixation paradigm.
Based on this dataset, we establish a benchmark for cognitive workload recognition with six workload levels and propose a learning framework that encodes spatiotemporal event representations.
Experimental results show that our approach achieves an average subject-specific accuracy of 96.36\% and 96.13\% under mixed random split evaluation.
These results suggest that event-based eye movements may provide a useful sensing pathway for future workload-aware rehabilitation and assistive interfaces.
\end{abstract}


\section{Introduction}
\label{sec:introduction}
\IEEEPARstart{S}{tate-aware} adaptation is a central requirement in modern rehabilitation and assistive interfaces~\cite{zhou2022cross}.
In stroke, traumatic brain injury, and other neurological conditions, therapy tasks must remain sufficiently challenging to promote engagement and learning, while excessive cognitive demand can reduce participation and training quality~\cite{gomes2025understanding}.
Prior work in robot-assisted gait rehabilitation has shown that cognitive load can be estimated and used to adapt the difficulty of virtual tasks for neurological patients in real time~\cite{koenig2011real}.
Similarly, Virtual Reality (VR)-based cognitive rehabilitation systems can adjust training difficulty to individual post-stroke cognitive deficits~\cite{maier2020adaptive}.
These studies highlight the need for unobtrusive cognitive-state monitoring methods that can support closed-loop adaptation in rehabilitation environments.

Eye movements provide a non-contact or minimally intrusive way for monitoring cognitive state~\cite{walter2021cognitive}.
Previous studies have shown that ocular behaviors such as fixation patterns, saccades, blinks, and microsaccades correlate with variations in mental effort~\cite{fadardi2022post, krejtz2018eye}.
Most existing eye-movement-based workload recognition methods rely on frame-based cameras or conventional eye trackers~\cite{de2009eye,aljehane2023studying,ktistakis2022colet}.
However, they are inherently limited by sampling rate, motion blur, and exposure constraints, which become particularly problematic under rapid eye movements, head-mounted configurations, and challenging illumination conditions commonly encountered in wearable rehabilitation and assistive systems.

Event cameras~\cite{dvs2008tobi}, also known as Dynamic Vision Sensors (DVS), offer an effective and efficient alternative for capturing fast ocular dynamics.
By capturing only brightness changes, they generate sparse asynchronous events with high temporal resolution, high dynamic range, and reduced motion blur.
These characteristics make event cameras suitable for recording rapid and subtle eye movements during target-guided interaction: they produce sparse data during fixation while still capturing fast eye movements during saccades.
Recent work has exploited event cameras for eye tracking~\cite{chen20233et,bonazzi2024retina,EGaze2024li,angelopoulos2020event,wang2024event,pei2024lightweight,zhao2024ev,li2023track,wang2024mambapupil,stoffregen2022event,lin2024fapnet,FACET2025}.
However, most studies focus on pupil center localization or gaze direction tracking, and rarely incorporate annotations or systematic evaluation of cognitive states such as workload.

Moreover, many prior workload recognition studies adopt free-viewing, visual-search, tracking, or weakly constrained interactive paradigms~\cite{ktistakis2022colet,miles2024cogload,nasri2024exploring,suslow2025look,chiossi2024understanding}, in which spatial gaze distributions and large-amplitude eye movements can provide strong discriminative cues for classification.
While effective in controlled laboratory settings, such paradigms may encourage models to rely on gaze distribution patterns rather than on workload-related ocular dynamics themselves.
Consequently, strong classification performance may not necessarily indicate that the model has captured genuine workload-related features, but instead that it has relied on task-specific patterns of gaze allocation that happen to be correlated with workload labels.

To address these limitations, this work investigates cognitive workload recognition from event-based eye movements under spatially constrained and task-driven conditions.
We introduce EveLoad, to the best of our knowledge, the first \underline{eve}nt-based eye-movement dataset for cognitive work\underline{load} recognition.
It is collected from 20 healthy participants under spatially constrained and task-driven conditions using a controlled N-back-guided~\cite{kirchner1958age,herff2014mental} fixation paradigm.
Although EveLoad is not a clinical rehabilitation dataset, it provides a preliminary exploration of event-based workload detection in controlled target-guided interaction, which is a relevant component for future workload-aware rehabilitation and assistive interfaces.
Based on this dataset, we develop a learning framework that encodes spatiotemporal event representations and evaluate multiple deep learning models under subject-specific and mixed-subject random split settings.

\begin{figure*}[t]
\centering
\includegraphics[width=\textwidth]{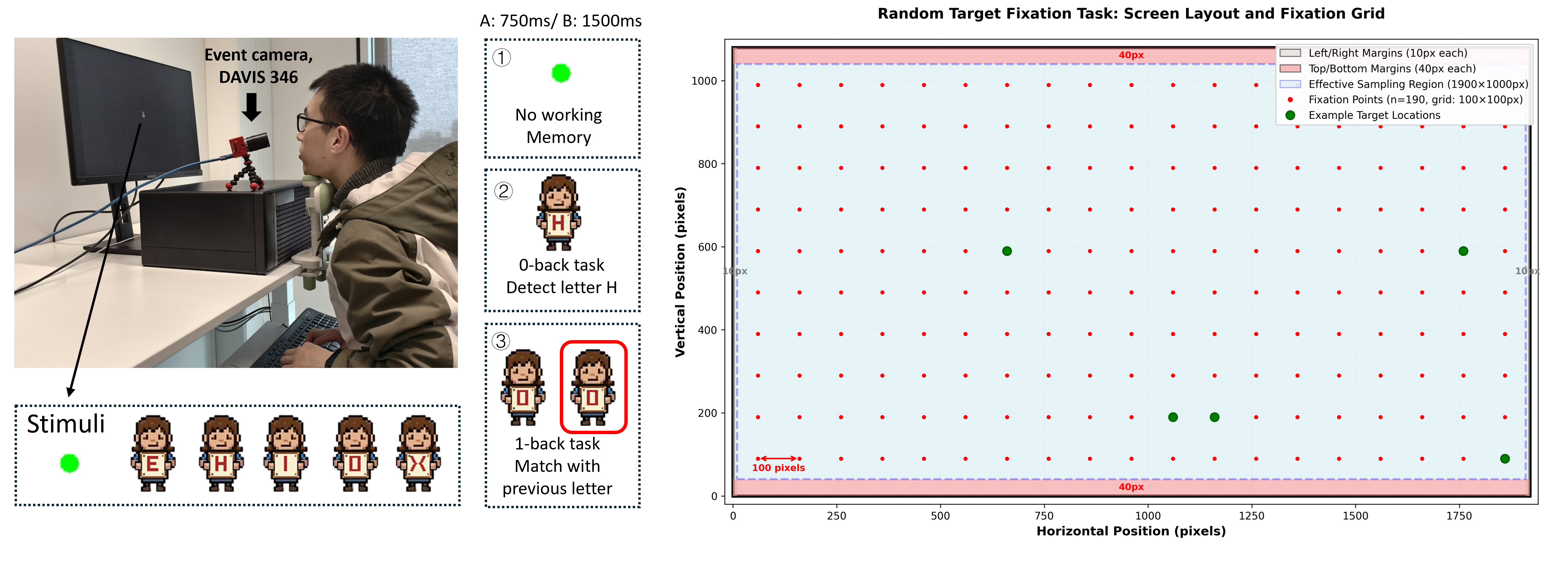}
\caption{Experimental setup and task paradigm for event-based cognitive workload recognition.
Left: Recording system with a DAVIS346 event camera capturing high-temporal-resolution eye movements during controlled fixation tasks. Middle: Stimulus design for no-load, 0-back, and 1-back working memory conditions with different presentation inter-stimulus interval (750\,ms, 1500\,ms). Right: Spatially constrained fixation grid and effective sampling region for the random target fixation task, ensuring consistent gaze behavior across stimuli.}
\label{fig:setup}
\end{figure*}

\section{Related Works}

This section reviews related work on event-based vision for eye-related sensing and cognitive workload monitoring for adaptive rehabilitation. 

\subsection{Event-based Vision for Eye Movement Sensing}

Event cameras are bio-inspired visual sensors that asynchronously report per-pixel brightness changes instead of recording dense frames at a fixed frame rate~\cite{dvs2008tobi,gallegoEventbasedVisionSurvey2022}.
Each event includes the spatial location, timestamp, and polarity of a local intensity change, producing a sparse spatiotemporal stream.
Compared with conventional frame-based cameras, event cameras provide high temporal resolution, high dynamic range, low latency, sparse output, and reduced motion blur, making them suitable for fast-motion and challenging-illumination scenarios.
These properties have motivated event-based methods for object tracking and high-speed human-motion capture~\cite{pan2019bringing,xu2020eventcap}.

These characteristics make event cameras well-suited for eye-movement sensing, because rapid saccades, blinks, pupil changes, and subtle ocular movements often occur on timescales that are difficult to capture reliably with low-frame-rate cameras.
Recent studies have explored event cameras for near-eye gaze tracking and pupil localization~\cite{angelopoulos2020event,stoffregen2022event,chen20233et,li2023track,EGaze2024li,zhao2024ev,bonazzi2024retina,wang2024event,pei2024lightweight,lin2024fapnet}, 
which demonstrate that event streams can support high-frequency and efficient eye tracking, but they mainly focus on gaze estimation, pupil-center localization, or eye-tracking algorithm design rather than eye-based cognitive-state recognition.
Very recent work has investigated cognitive load prediction using event-based human-pose estimation~\cite{aitsam2024measuring}, however, it does not provide an eye-movement cognitive workload benchmark.
Therefore, to the best of our knowledge, there is no event-based eye-movement dataset and benchmark for cognitive workload recognition.

\subsection{Cognitive Workload Monitoring for Adaptive Rehabilitation}

In rehabilitation engineering, cognitive workload is not only an assessment variable but also a potential control signal for adaptive therapy.
Koenig et al. demonstrated real-time closed-loop control of cognitive load in neurological patients during robot-assisted gait training, where a virtual task was adapted to avoid both overload and under-challenge~\cite{koenig2011real}.
VR-based cognitive rehabilitation has also adapted training to chronic stroke patients' individual cognitive deficits~\cite{maier2020adaptive}.
These studies show the clinical relevance of workload-aware adaptation and motivate continuous, unobtrusive monitoring methods.

Cognitive workload has been measured using subjective ratings, task-performance measures, physiological and electrophysiological signals, and ocular or vision-based features~\cite{kosch2023survey,tao2019systematic}.
Subjective and performance-based measures are useful for labeling and validation, but they are not continuous sensing streams.
Physiological signals such as electroencephalography (EEG)~\cite{zhang2019spectral}, functional near-infrared spectroscopy (fNIRS)~\cite{lim2020unified}, electrocardiography (ECG)~\cite{hogervorst2014combining}, heart-rate variability~\cite{john2022unraveling}, electrodermal activity, respiration~\cite{hogervorst2014combining}, and electrooculography (EOG)~\cite{belkhiria2021eog} provide objective information, but often require contact sensors, careful placement, calibration, or controlled acquisition conditions.
Their sensitivity to motion, contact instability, and environmental interference motivates more robust and unobtrusive modalities for interactive, assistive, and rehabilitation systems.

Ocular and vision-based sensing is attractive because eye movements are closely linked to attention, working memory, mental effort, and workload, while vision-based sensing can be non-contact, easy to deploy, and relatively unobtrusive.
Common indicators include pupil diameter, blink behavior, fixation duration, saccades, microsaccades, and smooth pursuit~\cite{johnston2022eyes,chen2022comparing,krejtz2018eye,fadardi2022post}.
Prior studies have explored these signals with eye trackers, infrared systems, RGB cameras, and headset-integrated sensors, including adaptive automation~\cite{de2009eye}, COLET eye-tracking data~\cite{ktistakis2022colet}, RGB camera recordings with deep learning~\cite{miles2024cogload}, headset-based multimodal sensing in XR environments~\cite{nasri2024exploring,hou2025cognitive}, and camera-based assessment in manufacturing scenarios~\cite{vasta2025evaluating}.
Compared with electrode-based physiological sensing, these approaches avoid attaching electrodes to the user and can be integrated into cameras, eye trackers, or head-mounted displays.

Nevertheless, most existing vision-based approaches still depend on frame-based sensing, where infrared and RGB systems rely on dense image acquisition and can be affected by illumination changes, motion blur, sampling-rate limits, and wearable power or form-factor constraints~\cite{shyam2020retaining,pan2019bringing}.
Event cameras offer a complementary route by asynchronously capturing sparse brightness changes, enabling low-latency sensing of rapid ocular dynamics with reduced motion blur and high dynamic range.
However, the absence of event-based eye-movement datasets with graded cognitive-workload annotations has limited systematic investigation of this modality.
EveLoad addresses this gap by providing such recordings and a baseline spatiotemporal pipeline for fine-grained cognitive workload recognition.

\section{EveLoad Dataset}

This section describes the construction of EveLoad, including the acquisition setup, task protocol, workload labels, and data organization. The data were collected from healthy participants under graded workload conditions induced by memory demand and stimulus pacing. EveLoad serves as a preliminary dataset for exploring the feasibility of cognitive workload estimation from event-based eye movements.

\begin{figure*}[t]
\centering
\includegraphics[width=\textwidth]{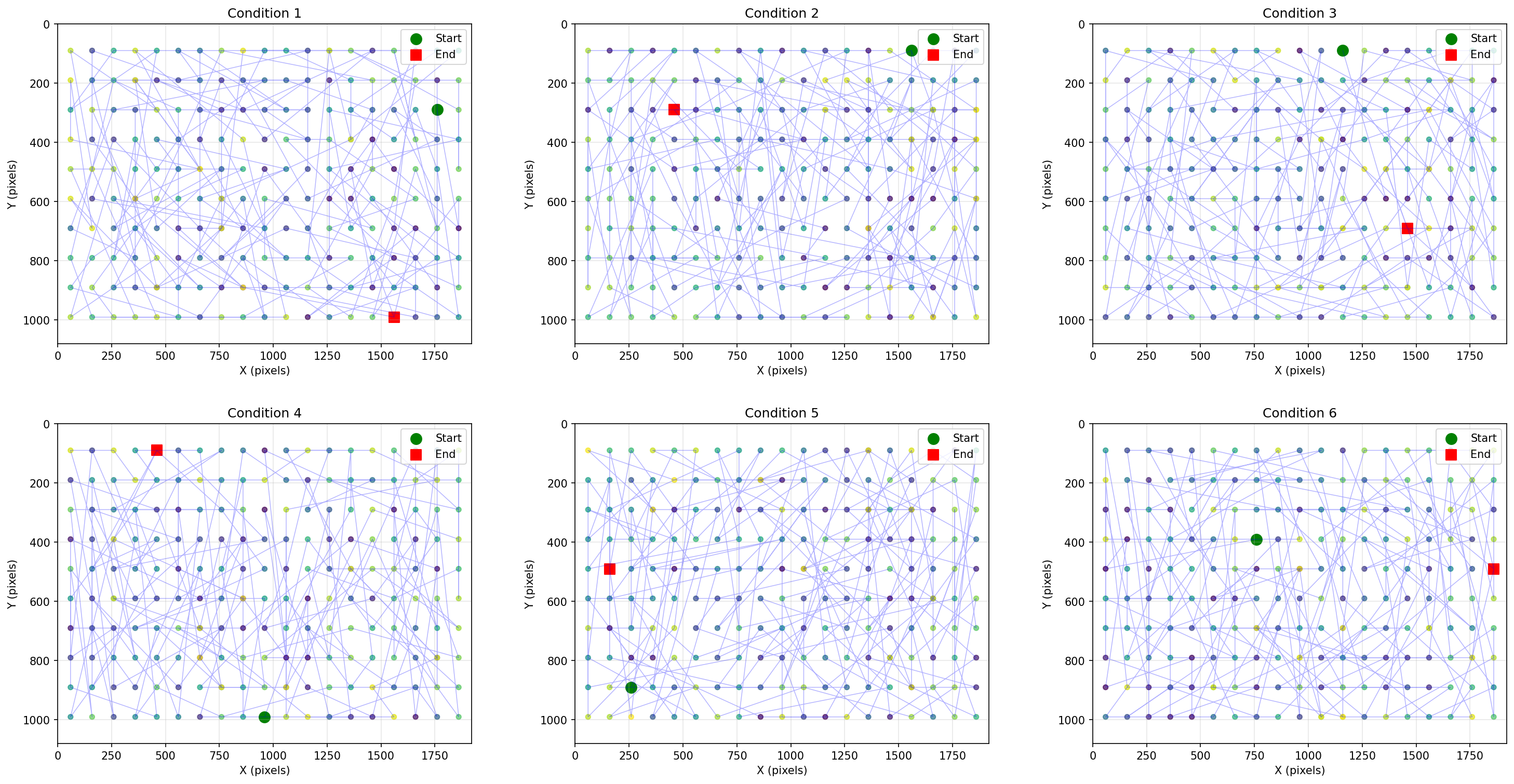}
\caption{Examples of constrained fixation trajectories under six experimental conditions. Each sequence consists of 200 target locations covering 190 predefined points within the effective sampling region. The displacement between consecutive targets is constrained to 200--800 pixels to ensure sufficient eye movement and uniform spatial coverage.}
\label{fig:trajectory}
\end{figure*}

\subsection{Data Acquisition Setup}
The dataset contains 7.5-hour recordings from 20 healthy adult participants (12 female, 8 male) with normal or corrected-to-normal vision, each performing a sequence of fixation tasks with varying cognitive demands.
As shown in Fig.~\ref{fig:setup}, during acquisition, participants sat approximately 40\,cm from a 1920 x 1080 display.
This setup was adopted to reduce head-motion artifacts and to ensure consistent eye-to-screen geometry across recordings.
All data were collected under natural ambient lighting conditions, which reflects realistic application scenarios.

Eye movements were recorded using an event-based vision sensor, DAVIS 346, positioned to capture eye activity. The sensor recorded asynchronous brightness changes with microsecond temporal resolution. Visual stimuli were presented on a front-facing monitor controlled by a dedicated experimental program. Stimulus presentation, keyboard responses, and event camera timestamps were synchronized through a unified logging system, enabling precise temporal alignment between visual events, eye movements, and participant responses.

In total, the dataset comprises synchronized event streams, fixation targets, behavioral responses, and cognitive workload annotations across 24{,}000 stimuli from 20 participants.
This count corresponds to 20 participants, six workload-speed levels, and 200 stimuli per level.

\subsection{Task Design}
Participants performed a continuous fixation task designed to constrain gaze behavior over the display area.
Specifically, a margin of 10 pixels is excluded from the left and right edges and 40 pixels from the top and bottom edges, resulting in an effective display region of approximately 1900 $\times$ 1000 pixels.
Within this region, target locations are sampled on a regular grid with a spacing of 100 $\times$ 100 pixels, yielding 190 fixed fixation points that are spatially uniform.
To reduce spatial target-location bias, we designed fixation target sequences with uniform coverage over the predefined sampling region.
Each sequence consisted of 200 target locations and covered all 190 predefined fixation points within the effective sampling region.
The displacement between consecutive targets was constrained to 200--800 pixels to ensure sufficient eye movement and avoid repetitive local patterns.
Fig.~\ref{fig:trajectory} illustrates example trajectories under six experimental conditions.

Regarding stimulus format, under the no-load working memory condition, the target was displayed as a green circular dot with a diameter of 15 pixels, as illustrated in Fig.~\ref{fig:setup}. This stimulus served solely to guide gaze shifts without imposing any perceptual or memory-related demands.
Under conditions with cognitive load, the fixation target was replaced by a pixelated stimulus region of approximately 100 x 100 pixels, containing an embedded letter stimulus that participants were required to evaluate. Each letter was rendered at approximately $15 \times 15$ pixels and was selected from the set \textbf{\{H, I, O, X, E\}}, as shown in Fig.~\ref{fig:setup}. This design ensured consistent visual saliency while enabling controlled manipulation of cognitive demand.
Cognitive workload was induced using 0-back and 1-back working-memory tasks.
In the 0-back condition, participants were instructed to judge whether the currently presented letter matched a predefined target letter (H). In the 1-back condition, participants judged whether the current letter was identical to the immediately preceding letter.
Participants responded using designated keyboard keys according to task instructions.
Participants responded via key presses: when the judgment condition was satisfied, they pressed the Enter key, otherwise, they pressed the Space key. 

\begin{table}[tb]
\centering
\caption{Experimental conditions and stimulus-task configurations.}
\label{tab:conditions}
\setlength{\tabcolsep}{1.2pt}
\begin{tabular}{l c c c c}
\hline
Level & Working Mem. & Stimulus & Decision Rule &
\begin{tabular}{c}
Inter-stimulus\\
Interval (ms)
\end{tabular} \\
\hline
L1 & --  & Green dot & Fixate on stimulus & 1500 \\
L2 & --  & Green dot & Fixate on stimulus & 750 \\
L3 & 0-back   & Letter patch & Match to letter H & 1500 \\
L4 & 0-back  & Letter patch & Match to letter H & 750 \\
L5 & 1-back  & Letter patch & Match to previous Letter & 1500 \\
L6 & 1-back  & Letter patch & Match to previous Letter & 750 \\
\hline
\end{tabular}
\end{table}

\begin{table}[tb]
\centering
\caption{Summary statistics of the EveLoad dataset.}
\label{tab:dataset_stats}
\setlength{\tabcolsep}{1pt}
\begin{tabular}{l c}
\hline
\textbf{Item} & \textbf{Description} \\
\hline
\#Subjects & 20 (12 female, 8 male) \\
Recording time per subj. & 22.5 min \\
Total recording time & 7.5 h \\
\#Stimuli & 24{,}000 \\
\#Level (\#Class) & 6 (No-load/0-back/1-back $\times$ Slow/Fast) \\
Annotation granularity & Stimulus-level \\
Labels per stimulus & Stimulus coordinate, workload level, response \\
\hline
\end{tabular}
\end{table}

Each task condition was implemented at two stimulus-pacing levels to manipulate time pressure while keeping the memory-load condition unchanged. 
The slow condition used a 1500-ms presentation duration, which is consistent with common timing settings in visual and verbal N-back tasks~\cite{chen2019testing, yeung2023changes}. 
The fast condition used a 750-ms presentation duration, motivated by prior sequential letter-based working-memory studies in which 750 ms and 1500 ms were used as adjacent presentation-time or presentation-rate levels~\cite{schmiedek2010hundred, schmiedek2014task}. 
Because manual alphabetic letter-identification responses are typically completed within approximately 0.5--0.6 s under forced-choice settings~\cite{mueller2012alphabetic}, the 750-ms condition was selected as a challenging but feasible response window rather than as a theoretical lower bound for visual recognition.

In total, as shown in Table~\ref{tab:conditions}, six experimental levels of cognitive workload were defined through the combination of memory load (no load, 0-back, 1-back) and presentation rates.
The combination of memory load and presentation rate provides a controllable workload ladder analogous to adaptive rehabilitation tasks, in which task complexity and pacing can be adjusted according to the user's current state.
The summary statistics of the EveLoad dataset can be found in Table~\ref{tab:dataset_stats}.

\begin{figure}[t]
    \centering
    \includegraphics[width=0.45\textwidth]{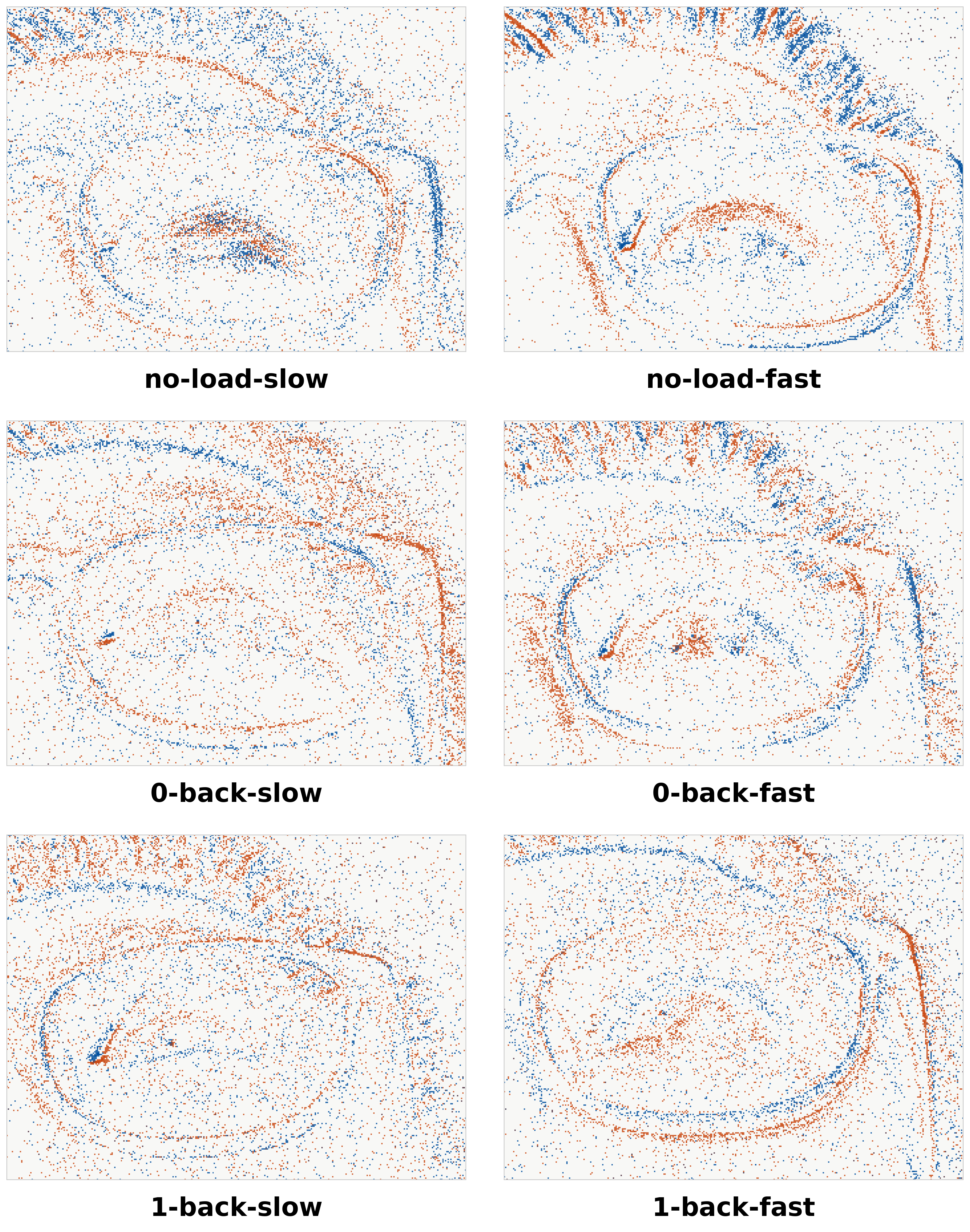}
    \caption{Examples of representation under different workload.}
    \label{fig:representation}
\end{figure}

\begin{figure*}[t]
    \centering
    \includegraphics[width=0.9\textwidth]{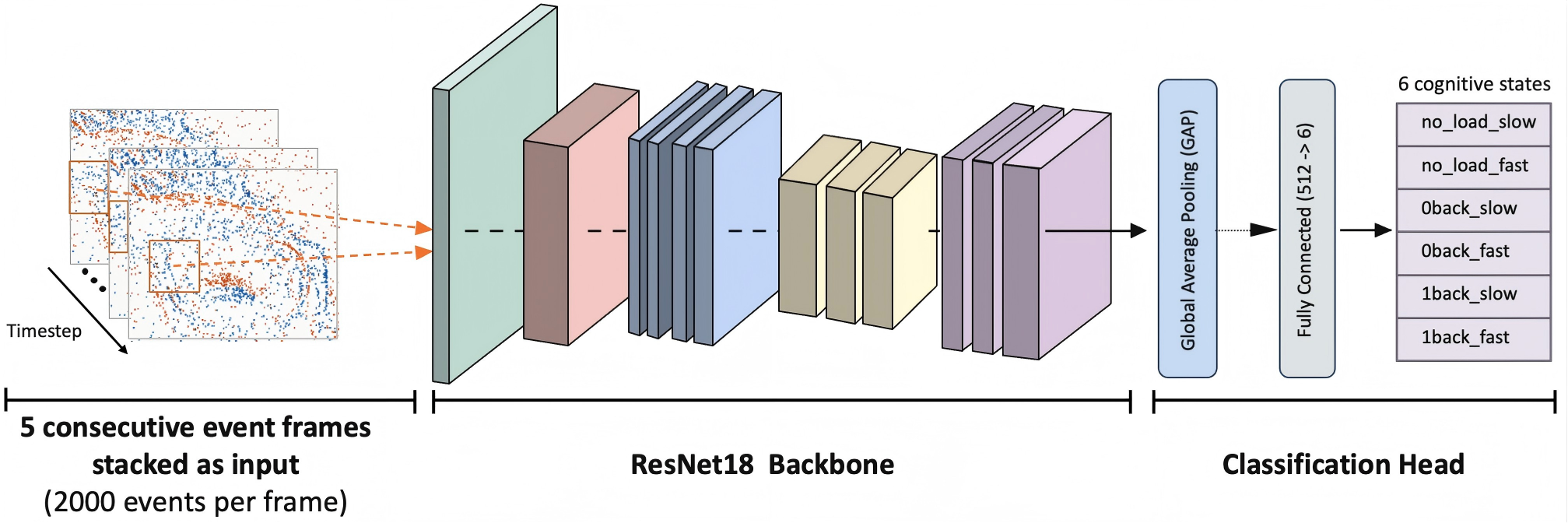}
    \caption{Overview of the pipeline for 6-class memory load prediction.
    The model takes five consecutive event frames, each accumulated from 2000 events, resulting in 10 polarity-stacked input channels. The feature extractor employs an adapted ResNet18 backbone initialized from ImageNet-pretrained weights. Global Average Pooling (GAP) is applied to the backbone feature maps, followed by a task-specific fully connected layer that maps the 512-dimensional feature vector to six workload logits.}
    \label{fig:workload_network_arch}
\end{figure*}

\refstepcounter{subsection}
\vspace{3.5ex plus 1.5ex minus 1.5ex}
\noindent{\color{subsectioncolor}\normalsize\normalfont\sf\itshape
\thesubsectiondis\hskip 0.5em Data Format\par}
\vspace{0.7ex plus .5ex minus 0ex}
\normalcolor
\normalcolor

All data streams were synchronized through a unified timestamping mechanism shared by the event camera and the experimental control system. Event streams, RGB images, stimulus events, and participant responses were recorded with microsecond-level timestamps, enabling precise temporal alignment between the fixation task and eye movement data.

Each fixation task was defined as an independent stimulus and annotated using a structured label file.
Annotations were organized at the stimulus level.
Each record includes the fixation target's screen coordinates $(x, y)$, the presented letter stimulus and its target response, the participant's response, and the stimulus start and end timestamps.
Participant and trial identifiers were also recorded.

Based on these annotations, each stimulus can be precisely aligned with a corresponding segment of event-based eye movement data and assigned a unique cognitive workload label.
This stimulus-level annotation scheme provides a unified and reproducible interface for subsequent gaze estimation and mental workload recognition analyses, and supports official stimulus-level split definitions for subject-specific and mixed random split evaluation.

\section{Methodology}
\normalcolor
This section introduces the methods used for cognitive workload classification, including data preprocessing (specifically hot-pixel removal and representation), the network model architecture, and the loss functions employed.

\subsection{Pre-processing}
\subsubsection{Hot Pixel Removal}
A hot pixel refers to a pixel location that continuously generates events at a fixed position. These pixels are typically spatially stationary after each device startup, and their number is usually fewer than five. However, hot pixels can significantly affect the generation of subsequent event representations. When a fixed event window is used, events generated by hot pixels may account for more than 90\% of the total events, severely reducing the proportion of valid events. In this study, we identified the coordinates of each hot pixel and filtered the raw event stream accordingly.

\subsubsection{Input Representation}
The model takes input as a tensor composed of a short sequence of event representation, also referred to as event frame. Each sample is represented as a four-dimensional tensor:
\begin{equation}
X \in \mathbb{R}^{B \times C \times H \times W},
\end{equation}
where $B$ denotes the batch size, $H \times W$ represents the spatial resolution of the event camera, and $C$ denotes the number of input channels.

Event frames are constructed using a fixed number of events per frame. Each frame is generated by accumulating 2000 consecutive events, ensuring a consistent amount of event information across different event rates. For each event, its spatial coordinate is mapped to the corresponding pixel location, and the value at that location is increased by one. Events are further separated according to polarity into negative and positive channels. Therefore, each event frame is represented as a two-channel event-count image, where the two channels correspond to the two polarities.

In this study, each input sample consists of 5 consecutive event frames stacked along the channel dimension. Since each frame contains two polarity channels, the resulting input representation comprises a total of 10 channels. This stacking strategy encodes short-term temporal dynamics into the channel dimension while preserving spatial structures.

\subsection{Neural Network Architecture}
\subsubsection{Backbone Architecture}
The cognitive workload recognition task adopts and compares three architectures---ResNet18~\cite{he2016deep}, MobileNetV3~\cite{howard2019searching}, and MobileViT~\cite{mehta2021mobilevit}---as feature extraction backbones. The network takes 5 consecutive polarity-stacked event frames as an input.
To enable the use of ImageNet-pretrained weights, the first convolutional layer is adapted to accommodate the 10-channel input. 
The backbone progressively extracts high-level representations while reducing spatial resolution. The resulting feature maps are processed by a global average pooling layer, which aggregates spatial information into a compact channel-wise feature vector.
\subsubsection{Classification Head}
For the final ResNet18 model, the adapted backbone extracts high-level spatial features, which are aggregated by global average pooling. The original ImageNet classifier is replaced with a task-specific fully connected layer that maps the 512-dimensional feature vector to six workload logits. For the MobileNetV3 and MobileViT baselines, we keep their original classifier structures and replace only the final classification layer for six-class prediction.

\subsection{Loss Function}

The cognitive load classification task is formulated as a six-class classification problem.
The model outputs prediction scores (logits) corresponding to the six cognitive load conditions.
Training is performed using the cross-entropy loss.

Let the model output for the $i$-th sample be $\mathbf{z}_i \in \mathbb{R}^{6}$ and the corresponding ground-truth label be $y_i \in \{0,\ldots,5\}$.
The classification loss is defined as:
\begin{equation}
\mathcal{L}_{\text{cls}}
= -\frac{1}{N}
\sum_{i=1}^{N}
\log
\frac{\exp(z_{i,y_i})}
{\sum_{c=1}^{6} \exp(z_{i,c})},
\end{equation}
where $N$ denotes the batch size and $z_{i,c}$ denotes the logit of the $i$-th sample for class $c$.

\section{Experiments}

This section evaluates the proposed EveLoad framework under subject-specific and mixed-subject random split settings.
We perform ablation studies on event representation, compare backbone architectures, analyze participant-wise performance, and compare EveLoad with representative eye-movement-based methods.
The goal is to assess whether event-based ocular signals contain workload-discriminative information under controlled target-guided interaction, which is a prerequisite for future workload-aware rehabilitation and assistive interfaces.

\begin{figure*}[t]
\centering
\includegraphics[width=\textwidth]{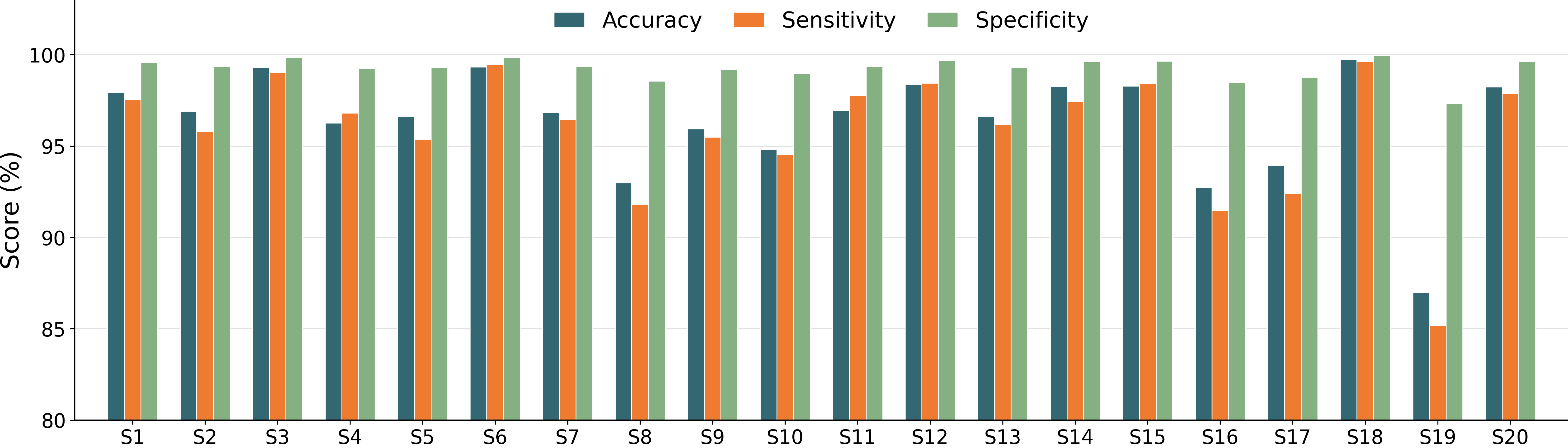}
\caption{Subject-specific classification performance of the final ResNet18 model on the test set for 20 participants. }
\label{fig:accuracy}
\end{figure*}

\begin{figure*}[t]
\centering
\includegraphics[width=\textwidth]{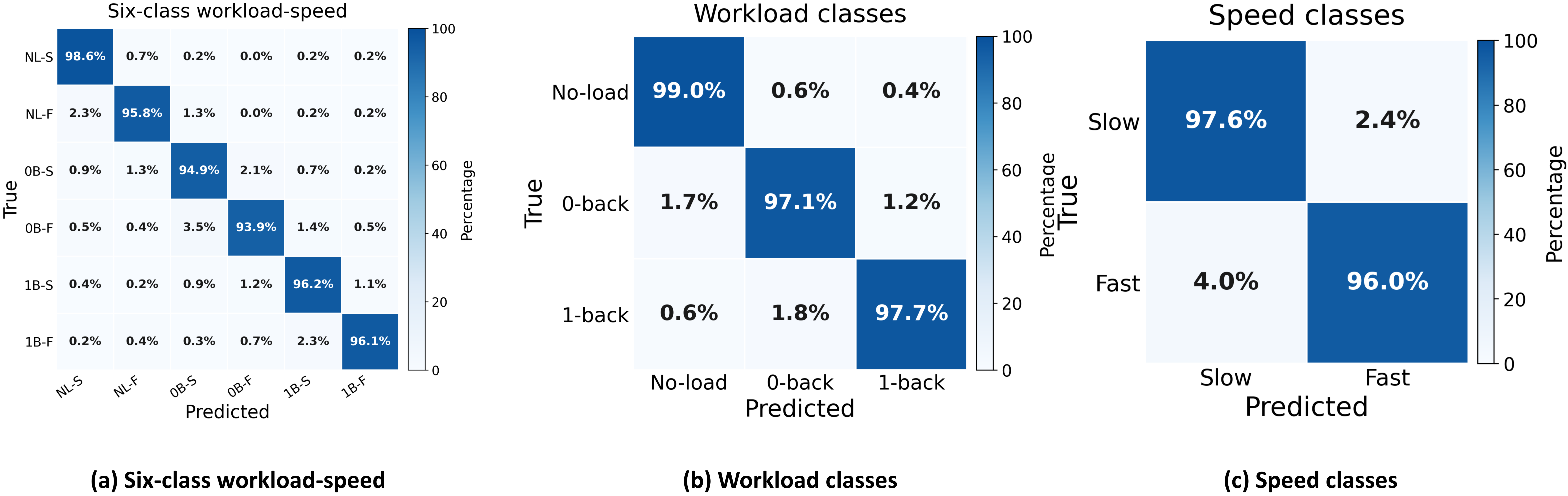}
\caption{Confusion matrices for EveLoad classification under the mixed random split. Rows denote true labels and columns denote predicted labels, and all values are row-normalized percentages. (a) Six-class workload-speed classification, with classes ordered by increasing workload level (no-load, 0-back, and 1-back) and, within each workload level, slow before fast. (b) Workload-level confusion matrix obtained by merging the slow and fast conditions within each workload level. (c) Speed-level confusion matrix obtained by merging workload levels into slow and fast conditions. }
\label{fig:mixed_split_confusion}
\end{figure*}

\subsection{Experimental Setup}

We evaluate our EveLoad framework under two different evaluation settings:

\begin{enumerate}
    \item A subject-specific setting is adopted, where models are trained and tested separately for each participant to investigate within-subject workload classification ability.
    Within each subject, stimuli are divided at the stimulus level into training, validation, and test sets with a ratio of 8:1:1, and stimuli from all conditions are then merged to form the final dataset.
    \item A mixed-subject random split is conducted at the stimulus level, where stimuli from all participants are merged and randomly divided into training, validation, and test sets with a ratio of 8:1:1.
    This setting evaluates mixed-subject stimulus-level recognition while still preserving the constraint that samples from the same stimulus do not cross split boundaries.
\end{enumerate}
Event frames are constructed directly from the raw event streams without additional spatial--temporal filtering or feature engineering.
Specifically, every 2000 events are accumulated to form one event frame, and 5 consecutive frames are stacked to generate a single input sample. Since event polarity is separated into two channels, each sample contains 10 input channels.
All reported results use a batch size of 32. Models were trained for 20 epochs with a fixed learning rate of $10^{-4}$ and a weight decay of $10^{-4}$.

\begin{table}[ht]
\centering
\caption{Ablation study of event frame construction using ResNet18 under subject-specific stimulus-level split. Each entry reports mean accuracy (\%) $\pm$ population standard deviation over 20 subjects.}
\label{tab:ablation_resnet18}
\footnotesize
\setlength{\tabcolsep}{3pt}
\renewcommand{\arraystretch}{1.05}
\begin{tabular}{@{}ccccc@{}}
\hline
Events/frame & $T=1$ & $T=5$ & $T=10$ & $T=15$ \\
\hline
500  & 89.79$\pm$5.27 & 94.68$\pm$3.87 & 95.27$\pm$3.63 & 95.46$\pm$4.37 \\
1000 & 92.38$\pm$4.44 & 95.73$\pm$3.04 & 96.18$\pm$3.09 & 95.74$\pm$3.87 \\
2000 & 94.12$\pm$4.18 & \textbf{96.36$\pm$2.90} & 95.34$\pm$4.45 & 94.64$\pm$4.78 \\
5000 & 95.55$\pm$3.78 & 95.11$\pm$4.22 & 94.11$\pm$5.65 & 92.29$\pm$5.64 \\
\hline
\end{tabular}
\end{table}
\begin{table}[ht]
\centering
\caption{Backbone comparison under the final input setting $(E=2000,T=5)$ with batch size 32. Metrics are reported under subject-specific stimulus-level split.}
\label{tab:model_comparison}
\resizebox{\columnwidth}{!}{
\begin{tabular}{lccc}
\hline
Model & Notes & Acc. (\%) & Std. (\%) \\
\hline
\textbf{ResNet18} & ImageNet-pretrained CNN, 11.20M & \textbf{96.36} & \textbf{2.90} \\
MobileViT & CNN-transformer hybrid, 4.94M & 94.12 & 4.64 \\
MobileNetV3 & Lightweight CNN, 4.21M & 92.21 & 6.09 \\
\hline
\end{tabular}
}
\end{table}

\begin{table*}[t]
\centering
\caption{Comparison of vision- and eye-movement-based cognitive workload recognition studies.}
\label{tab:comparison}

{
\setlength{\tabcolsep}{1.55pt}
\renewcommand{\arraystretch}{1.05}
\small

\def\evcell#1{\begin{tabular}[c]{@{}l@{}}#1\end{tabular}}
\def\evcellc#1{\begin{tabular}[c]{@{}c@{}}#1\end{tabular}}
\def\evitemgap{\noalign{\vskip 2.2mm}}

\begin{tabular*}{\textwidth}{@{\extracolsep{\fill}}lccclllllc@{}}
\hline
Study & Dataset & \#Subj. & \#Class & Task & Device & Features & Method & Split Setting & Acc. (\%) \\
\hline

\evcell{Skaramagkas et al.~\cite{Skaramagkas2021BIBE}\\(2021)}
& \evcellc{Private}
& \evcellc{37}
& \evcellc{2/3}
& \evcell{Visual\\Search}
& \evcell{Pupil\\Core}
& \evcell{Hand-crafted}
& \evcell{RF,\\SVM}
& \evcell{Random split\\(8:2)}
& \evcellc{88.0} \\
\evitemgap

\evcell{Ktistakis et al.~\cite{ktistakis2022colet}\\(2022)}
& \evcellc{COLET}
& \evcellc{47}
& \evcellc{4}
& \evcell{Visual\\Search}
& \evcell{Pupil\\Core}
& \evcell{Statistical}
& \evcell{Ensemble}
& \evcell{Random split\\(8:2)}
& \evcellc{88.0} \\
\evitemgap

\evcell{Miles et al.~\cite{miles2024cogload}\\(2024)}
& \evcellc{EM-\\COGLOAD}
& \evcellc{75}
& \evcellc{2}
& \evcell{Tracking}
& \evcell{Standard\\Camera}
& \evcell{Time-series}
& \evcell{CNN}
& \evcell{Subject-\\independent}
& \evcellc{87.5} \\
\evitemgap

\evcell{Nasri et al.~\cite{nasri2024exploring}\\(2024)}
& \evcellc{Private}
& \evcellc{19}
& \evcellc{2}
& \evcell{VR\\Training}
& \evcell{Varjo\\VR-3}
& \evcell{Hand-crafted}
& \evcell{MLP,\\RF}
& \evcellc{--}
& \evcellc{84.0} \\
\evitemgap

\evcell{Hou et al.~\cite{hou2025cognitive}\\(2025)}
& \evcellc{Private}
& \evcellc{20}
& \evcellc{3}
& \evcell{MR CNC\\Operation}
& \evcell{HoloLens\\2}
& \evcell{Head, eye,\\hand}
& \evcell{Transformer-\\CL}
& \evcell{Validation split\\(8:2)}
& \evcellc{95.83} \\
\evitemgap


\evcell{Wibirama et al.~\cite{wibirama2026cognitive}\\(2026)}
& \evcellc{Private}
& \evcellc{40}
& \evcellc{2}
& \evcell{Online\\Shopping}
& \evcell{Gazepoint\\GP3}
& \evcell{Raw gaze\\series}
& \evcell{ALSTM-\\FCN}
& \evcell{Participant\\walk-forward}
& \evcellc{97.70} \\
\evitemgap

\evcell{Shao et al.~\cite{shao2026multimodal}\\(2026)}
& \evcellc{Private}
& \evcellc{26}
& \evcellc{2}
& \evcell{Driving\\Simulator}
& \evcell{Dikablis\\Glasses 3}
& \evcell{Eye, physio,\\driving}
& \evcell{SVM}
& \evcell{Validation split\\(8:2)}
& \evcellc{87.8} \\
\evitemgap

\evcell{\textbf{Ours}\\\textbf{(EveLoad)}}
& \evcellc{EveLoad}
& \evcellc{20}
& \evcellc{6}
& \evcell{Constrained\\Fixation}
& \evcell{Event\\Camera}
& \evcell{Spatio-\\temporal}
& \evcell{ResNet18}
& \evcell{Subject-specific / mixed\\random split (8:1:1)}
& \evcellc{\textbf{96.36 /}\\\textbf{96.13}} \\

\hline
\end{tabular*}

\vspace{1mm}
\begin{minipage}{\textwidth}
\scriptsize
\raggedright
Note: Reported accuracies follow the cited papers and are not directly comparable because datasets, class numbers, label definitions, sensors, and validation protocols differ.
\end{minipage}
}

\end{table*}

\subsection{Ablation Study}
We first evaluate the effect of event representation construction using ResNet18 under the subject-specific setting. Table~\ref{tab:ablation_resnet18} reports the 4 $\times$ 4 grid over events per frame and the number of consecutive frames.

The best performance is obtained with 2000 events per frame and 5 consecutive frames, achieving an average subject-specific accuracy of 96.36\% with a standard deviation of 2.90\%. This setting also yields the lowest standard deviation among the top-performing configurations and is therefore used as the final input setting in subsequent experiments.

\refstepcounter{subsection}
\vspace{3.5ex plus 1.5ex minus 1.5ex}
\noindent{\color{subsectioncolor}\normalsize\normalfont\sf\itshape
\thesubsectiondis\hskip 0.5em Model Comparison\par}
\vspace{0.7ex plus .5ex minus 0ex}
\normalcolor
Table~\ref{tab:model_comparison} compares ResNet18, MobileViT, and MobileNetV3 under the selected input setting. ResNet18 achieves the highest average accuracy of 96.36\%, outperforming MobileViT and MobileNetV3 by 2.24 and 4.15 percentage points, respectively. This result indicates that a moderately sized convolutional backbone provides a stronger representation for event-based workload recognition than the lighter MobileNetV3 and the MobileViT variant in this dataset.

\subsection{Performance Analysis}
Fig.~\ref{fig:accuracy} presents the subject-specific classification accuracy on the test set for all 20 participants.
The final ResNet18 model achieves consistently high within-subject performance for most participants, with a user-level mean accuracy of 96.36\% and a macro F1 score of 96.00\%.
The mixed-subject random split achieves a comparable accuracy of 96.13\% and a macro F1 score of 95.84\%, suggesting that the discriminative structure of the six workload classes remains stable when stimuli from all subjects are jointly modeled at the stimulus level.
These results indicate that event-based ocular signals contain workload-discriminative information under controlled target-guided interaction, which is a prerequisite for future closed-loop rehabilitation interfaces.


\subsection{Comparison analysis}
Table~\ref{tab:comparison} compares EveLoad with representative vision- and eye-movement-based cognitive workload recognition studies, including studies based on conventional eye trackers, RGB cameras, Mixed Reality Head-Mounted Display (MR-HMD) sensors, and multimodal driver-monitoring signals~\cite{Skaramagkas2021BIBE,ktistakis2022colet,miles2024cogload,nasri2024exploring,hou2025cognitive,dell2025your,wibirama2026cognitive,shao2026multimodal}. A key distinction lies in the control of target-location distributions during data collection. Existing studies are commonly conducted in free-viewing, visual-search, visual-tracking, driving, online-shopping, or complex interactive scenarios, where gaze locations are largely influenced by visual content, target layout, task trajectory, road context, or interaction process. Although such paradigms are effective for inducing cognitive workload variations, they may allow classifiers to exploit workload-related differences in spatial gaze distributions, rather than relying solely on ocular dynamics. In contrast, EveLoad adopts a spatially constrained fixation paradigm in which target locations are sampled from a predefined grid and each workload condition covers the same sampling region. As illustrated in Fig.~\ref{fig:trajectory}, the marginal spatial distribution of fixation targets remains approximately uniform across workload conditions. This design reduces the possibility that workload labels can be inferred merely from target-location density.

Another important difference is the granularity of workload annotation. Most prior studies formulate cognitive workload recognition as a binary classification problem, while even the most fine-grained setting among the compared works contains four workload-related classes. Such settings provide useful but relatively coarse estimates of mental workload. In contrast, EveLoad defines six workload levels by combining memory load and stimulus presentation rate. This finer label granularity provides a more detailed benchmark for cognitive workload recognition and has the potential to better reflect subtle variations in mental demand, which is particularly important for adaptive rehabilitation and assistive interfaces that require awareness of users' cognitive states.

Finally, to the best of our knowledge, EveLoad is the first cognitive workload dataset collected using event-based eye-movement sensing.
Unlike conventional frame-based eye trackers, event cameras asynchronously capture brightness changes with high temporal resolution, low latency, and reduced motion blur, making them suitable for recording rapid and subtle ocular dynamics.
As shown in Table~\ref{tab:comparison}, our method achieves 96.36\% accuracy under subject-specific evaluation and 96.13\% accuracy under the mixed random stimulus-level split.
Although results across different datasets and task designs should not be interpreted as a strictly controlled cross-dataset comparison, the strong recognition performance obtained on a six-class workload benchmark provides preliminary evidence for the potential of event-based ocular sensing in fine-grained cognitive workload monitoring under spatially constrained task conditions.
This sensing pathway may complement physiological sensors and frame-based eye tracking in future workload-aware adaptive rehabilitation and assistive interfaces.

\subsection{Implications for Adaptive Rehabilitation Interfaces}

From a rehabilitation-engineering perspective, EveLoad should be viewed as an enabling benchmark for workload-aware adaptive interfaces rather than as a clinical outcome study. In closed-loop rehabilitation, estimates of cognitive state can serve as a real-time control signal to adjust task difficulty, stimulus pacing, feedback intensity, therapist alerts, rest-break scheduling, or adaptive AR/VR cognitive-motor training. Such adaptation may help keep training challenging and engaging without overloading the patient or increasing fatigue. Event-based eye-movement sensing is attractive for this purpose because it can capture rapid ocular dynamics with high temporal resolution and low latency, and may be integrated into future head-mounted rehabilitation systems.

The present results show that such signals have the potential to support fine-grained workload recognition in healthy participants under controlled task conditions. These findings provide an initial technical basis for using event-based ocular signals in adaptive rehabilitation settings. Future work should validate this approach in stroke, traumatic brain injury, and other neurological populations, where cognitive workload and eye-movement patterns may differ from those of healthy users. Further studies should also evaluate whether online event-based workload estimates can improve adaptive rehabilitation protocols and lead to better training engagement, tolerance, or functional outcomes.

\section{Conclusion}
In this paper, we introduced EveLoad, an event-based eye-movement dataset with graded cognitive workload annotations collected from 20 healthy participants under spatially constrained and task-driven conditions.
The dataset provides synchronized event streams, behavioral responses, and stimulus-level cognitive workload labels, enabling systematic evaluation of workload recognition models under controlled target-guided interaction.
Based on EveLoad, we established a benchmark for cognitive workload recognition using deep learning models.
Experimental results demonstrate that event-based eye movements contain informative cues for workload discrimination, enabling high recognition accuracy under subject-specific and mixed random split settings.
These findings support the potential of event-based eye-movement sensing as a building block for future workload-aware rehabilitation and assistive technologies.
Clinical deployment will further require validation in stroke, traumatic brain injury, and other neurological populations, as well as integration with real-time therapy systems.

\bibliographystyle{IEEEtran}
\bibliography{ref}

\end{document}